%% file: main.tex
\documentclass[10pt,twocolumn,letterpaper]{article}

\usepackage[pagenumbers]{cvpr} 

\input{preamble}

\def\paperID{5266} 
\def\confName{CVPR}
\def\confYear{2024}

\title{Towards a Perceptual Evaluation Framework for Lighting Estimation}

\author{Justine Giroux$^1$ \quad
Mohammad Reza Karimi Dastjerdi$^1$ \quad
Yannick Hold-Geoffroy$^2$ \\
Javier Vazquez-Corral$^{3,4}$ \quad
Jean-Fran\c{c}ois Lalonde$^1$ \\
$^1$Université Laval, $^2$Adobe Research, $^3$Computer Vision Center, $^4$Universitat Autònoma de Barcelona
}

\begin{document}
\twocolumn[{%
\renewcommand\twocolumn[1][]{#1}%

\maketitle
\vspace{-10mm}
\begin{center}
    \centering
    \captionsetup{type=figure}
    \includegraphics[width=1.0\linewidth]{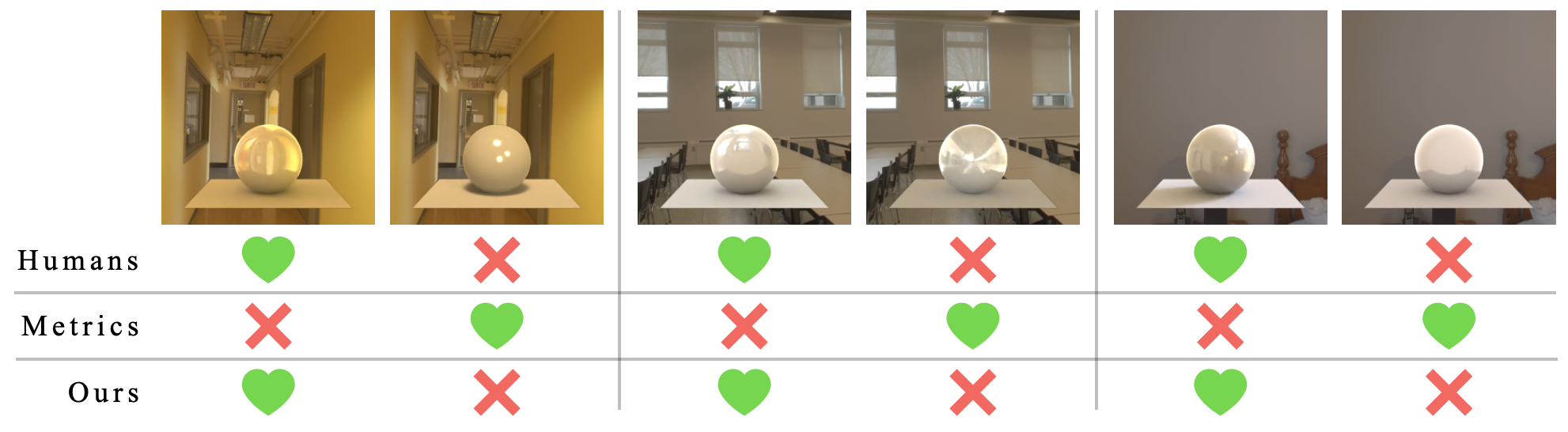}
    \captionof{figure}{We pit image comparison metrics, used to quantify the performance of lighting estimation algorithms, against human perception.  
    When asked which render looks most plausible, our controlled psychophysical study reveals that humans preference contradicts image metrics in the vast majority of cases. This paper questions the current practice of employing image quality metrics for evaluating lighting estimation algorithms when used for the task of virtual object insertion: can we do better by considering human perception? 
    \vspace{0.2cm}
   }
   \label{fig:teaser}
\end{center}%
}]

\input{sec/0_abstract}    
\input{sec/1_intro}

\input{sec/2_related_work}
\input{sec/3_psychophys_exp}

\input{sec/4_results}
\input{sec/5_measure_comparison}
\input{sec/6_metric}
\input{sec/7_conc}

%

{
    \small
    \bibliographystyle{ieeenat_fullname}
    \bibliography{main}
}


\end{document}

%% file: preamble.tex
\usepackage[dvipsnames]{xcolor}

\usepackage{microtype}
\usepackage{siunitx}
\usepackage{physics}
\usepackage{todonotes}
\usepackage{tabularx}
\usepackage{longtable}
\usepackage[accsupp]{axessibility} 

\makeatletter
\newbox\LT@firstfoot
\def\endfirstfoot{\LT@end@hd@ft\LT@firstfoot}
\newdimen\LT@footdiff
\def\LT@start{%
  \let\LT@start\endgraf
  \endgraf\penalty\z@
  \vskip\LTpre\endgraf
  \LT@footdiff-\ht\LT@foot
  \advance\LT@footdiff\ht\LT@firstfoot
  \dimen@\pagetotal
  \advance\dimen@ \ht\ifvoid\LT@firsthead\LT@head\else\LT@firsthead\fi
  \advance\dimen@ \dp\ifvoid\LT@firsthead\LT@head\else\LT@firsthead\fi
  \advance\dimen@ \ht\ifvoid\LT@firstfoot\LT@foot\else\LT@firstfoot\fi
  \dimen@ii\vfuzz
  \vfuzz\maxdimen
  \setbox\tw@\copy\z@
  \setbox\tw@\vsplit\tw@ to \ht\@arstrutbox
  \setbox\tw@\vbox{\unvbox\tw@}%
  \vfuzz\dimen@ii
  \advance\dimen@ \ht
      \ifdim\ht\@arstrutbox>\ht\tw@\@arstrutbox\else\tw@\fi
  \advance\dimen@\dp
      \ifdim\dp\@arstrutbox>\dp\tw@\@arstrutbox\else\tw@\fi
  \advance\dimen@ -\pagegoal
  \ifdim \dimen@>\z@\vfil\break\fi
  \global\@colroom\@colht
  \ifvoid\LT@firstfoot
    \ifvoid\LT@foot
    \else
      \advance\vsize-\ht\LT@foot
      \global\advance\@colroom-\ht\LT@foot
      \dimen@\pagegoal\advance\dimen@-\ht\LT@foot\pagegoal\dimen@
      \maxdepth\z@
    \fi
  \else
    \advance\vsize-\ht\LT@firstfoot
    \global\advance\@colroom-\ht\LT@firstfoot
    \dimen@\pagegoal\advance\dimen@-\ht\LT@firstfoot\pagegoal\dimen@
    \maxdepth\z@
  \fi
  \ifvoid\LT@firsthead\copy\LT@head\else\box\LT@firsthead\fi\nobreak
  \output{\LT@output}%
}
\def\LT@output{%
  \ifnum\outputpenalty <-\@Mi
    \ifnum\outputpenalty > -\LT@end@pen
      \LT@err{floats and marginpars not allowed in a longtable}\@ehc
    \else
      \setbox\z@\vbox{\unvbox\@cclv}%
      \ifdim \ht\LT@lastfoot>\ht\LT@foot
        \dimen@\pagegoal
        \advance\dimen@-\ht\LT@lastfoot
        \ifdim\dimen@<\ht\z@
          \setbox\@cclv\vbox{\unvbox\z@\copy\LT@foot\vss}%
          \@makecol
          \@outputpage
          \setbox\z@\vbox{\box\LT@head}%
        \fi
      \fi  
      \global\@colroom\@colht
      \global\vsize\@colht   
      \vbox
        {\unvbox\z@\box\ifvoid\LT@lastfoot\LT@foot\else\LT@lastfoot\fi}%
    \fi
  \else
    \ifvoid\LT@firstfoot
      \setbox\@cclv\vbox{\unvbox\@cclv\copy\LT@foot\vss}%
      \@makecol
      \@outputpage
      \global\vsize\@colroom
    \else
      \setbox\@cclv\vbox{\unvbox\@cclv\box\LT@firstfoot\vss}%
      \@makecol
      \@outputpage
      \global\advance\@colroom\LT@footdiff
      \global\vsize\@colroom
    \fi
    \copy\LT@head\nobreak
  \fi
}
\makeatother

\makeatletter 
\let\c@table\c@figure
\let\c@lstlisting\c@figure
\makeatother

\DeclareSIUnit{\px}{px}

\definecolor{everlight}{rgb}{0.7490196078431373, 0.3176470588235294, 0.23921568627450981}
\definecolor{weber22}{rgb}{0.788235294117647, 0.6352941176470588, 0.5882352941176471}
\definecolor{gardner19_3}{rgb}{0.7725490196078432, 0.7568627450980392, 0.5333333333333333}
\definecolor{stylelight}{rgb}{0.4823529411764706, 0.6941176470588235, 0.5686274509803921}
\definecolor{average_image_texture}{rgb}{0.8509803921568627, 0.6823529411764706, 0.18823529411764706}
\definecolor{everlight_outdoor}{rgb}{0.7490196078431373, 0.3176470588235294, 0.23921568627450981}
\definecolor{jinsong}{rgb}{0.7725490196078432, 0.7568627450980392, 0.5333333333333333}
\definecolor{average_image_outdoor_texture}{rgb}{0.8509803921568627, 0.6823529411764706, 0.18823529411764706}

\definecolor{cvprblue}{rgb}{0.21,0.49,0.74}
\usepackage[pagebackref,breaklinks,colorlinks,citecolor=cvprblue]{hyperref}

\newcommand{\valeur}[1]{#1}

\setuptodonotes{inline, color=JungleGreen!30, bordercolor=white}

\newcommand{\myparagraph}[1]{\noindent \textbf{#1} \thickspace}

%% file: sec/0_abstract.tex
\begin{abstract}
Progress in lighting estimation is tracked by computing existing image quality assessment (IQA) metrics on images from standard datasets. While this may appear to be a reasonable approach, we demonstrate that doing so does not correlate to human preference when the estimated lighting is used to relight a virtual scene into a real photograph. To study this, we design a controlled psychophysical experiment where human observers must choose their preference amongst rendered scenes lit using a set of lighting estimation algorithms selected from the recent literature, and use it to analyse how these algorithms perform according to human perception. Then, we demonstrate that none of the most popular IQA metrics from the literature, taken individually, correctly represent human perception. Finally, we show that by learning a combination of existing IQA metrics, we can more accurately represent human preference. This provides a new perceptual framework to help evaluate future lighting estimation algorithms. 
To encourage future research, all (anonymised) perceptual data and code are available at {\small\url{https://lvsn.github.io/PerceptionMetric/}}.
\end{abstract}


%% file: sec/1_intro.tex
\section{Introduction}
\label{sec:intro}

Light fashions the appearance of everything we see. From the warm and glowing golden hour to the coldness of neon tubes, lighting determines how scenes look and how we perceive them~\cite{murray2019visual}. This wide variety in illumination conditions creates a challenge for computer vision algorithms attempting to estimate lighting from images. How can it be robustly estimated from images when conditions are so different? Luckily, large lighting datasets~\cite{gardner2017learning,bolduc2023beyondthepixel} and learning-based approaches~\cite{hold2017deep,cheng2018shlight,gardner2017learning} have enabled significant progress over the past decade. Nowadays, automatic lighting estimation methods allow artists to plausibly relight virtual objects and composite them into real images with ever-increasing ease.


As with many computer vision tasks, progress in lighting estimation is tracked by computing existing image quality assessment (IQA) metrics on standard datasets. Since lighting representations differ (e.g., environment maps~\cite{gardner2017learning}, parametric lights~\cite{gardner2019deep,zhang2019all}, spherical harmonics~\cite{garon2019fast}, spherical Gaussians~\cite{li2020inverse}, etc.), it is common practice to render a virtual object with the estimated lighting, and compare it with a render of the same virtual object lit with the ground truth. A variety of existing IQA metrics---ranging from classical approaches (e.g., RMSE and SSIM~\cite{SSIM}) to those designed to mimic certain aspects of human perception (e.g., LPIPS~\cite{LPIPS})---have been used for this purpose. 

Whilst this way of quantifying performance appears reasonable, it is unclear whether doing so follows human preference. After all, once a virtual object is relit and inserted into an image, perfectly matching ground truth lighting may not be needed to achieve plausible results. In a practical scenario, one does not have access to the ground truth and must judge the realism of a relighting result solely based on its appearance. 
For example, consider the images shown in \cref{fig:teaser}. When asked to choose which of the two images looks most realistic, humans prefer the left one. However, IQA metrics establish that the right one most closely matches the ground truth (not shown). In this work, we question the validity of using IQA metrics for evaluating lighting estimation algorithms: beyond the selected examples of \cref{fig:teaser}, do IQA metrics generalise and agree with human perception? Or is another approach necessary? 




In contrast to common belief, we show that this strategy of employing existing IQA metrics for determining whether an approach outperforms previous work \emph{does not reflect human preference}. To do so, we perform a controlled psychophysical study where a set of lighting estimation approaches, selected from the recent literature, are pitted against each other by asking observers to judge the accuracy and plausibility of virtual object insertion results. 
Our perceptual data reveals that no single IQA metric agrees with human perception across all situations. 
%
Finally, we show that a learned combination of the same IQA metrics, trained on our perceptual data, achieves results that match human preference. 




We summarise our contributions as follows. First, we perform a psychophysical study comparing recent lighting estimation methods by asking observers to judge scenes rendered with lighting estimates. Second, we evaluate the agreement between commonly-used IQA metrics and our perceptual data. In particular, we find that most IQA metrics do not corroborate human perception. Third, instead of considering IQA metrics individually, we show that by \emph{combining} them in a learned fashion, we can more accurately mimic human perception for the evaluation of illumination conditions. This provides a new framework for evaluating future illumination estimation algorithms in a way that more closely matches human perception.


%% file: sec/2_related_work.tex
\section{Related work}
\label{sec:related_work}

\paragraph{Lighting estimation.}
\label{subsec:related_work_lighting_estimation_methods}
Outdoor illumination estimation traditionally relied on manual identification of features such as sky, ground shadows, and surface shading~\cite{lalonde2012estimating}, or optimising lighting and reflectance interchangeably~\cite{lalonde2014lighting}. Deep learning replaced these manual features with implicitly learned representations, where two predominant categories emerge. The first category utilises parametric representations for illumination estimation, such as spherical harmonics~\cite{garon2019fast} and spherical Gaussians~\cite{gardner2019deep,li2020inverse,wang2021learning,li2023spatiotemporally} for indoor, and sky models for outdoor scenarios \cite{hold2017deep,zhang2019all}. While these methods are easily editable, they often fail to produce a plausible environment, limiting its realism. The second category encompasses environment map prediction models \cite{gardner2017learning,legendre2019deeplight,wang2022stylelight}. These models predict the complete scene field of view as a \ang{360} environment map. While yielding high-quality and realistic outputs, these works are complex to train and lack editability. Recent research has sought to integrate these categories~\cite{weber2022editable,Dastjerdi_2023_ICCV}, employing a two-stage approach which first predicts a parametric lighting representation and then uses it to condition environment map generation. 

\input{sec/figs_latex/tex_fig_exp_tasks}
\myparagraph{Image quality assessment (IQA) metrics.}
\label{subsec:related_work_IQA_metrics}
IQA has been an ongoing research topic for more than 20 years \cite{SSIM}. It particularly flourished with the appearance of mobile phone cameras and their ubiquitous presence in our current daily lives. IQA metrics can be divided into three types: Full-Reference, Reduced-Reference, and No-Reference. Whilst Reduced-Reference metrics such as FID \cite{fid} and Inception Score~\cite{IS}, or No-Reference-based metrics such as BRISQUE~\cite{BRISQUE}, NIQE~\cite{NIQE}, UNIQUE~\cite{zhang2021uncertainty}, HyperIQA~\cite{HyperIQA}, and \cite{freitas2020image} show promising applications, the interest of our work lies in comparing images using a Reference-based metric. Domain-specific IQA metrics were also proposed, such as DISTS \cite{ding2020image,ding2021locally} for textures; in the remainder of the document, we will focus on generic scene-level metrics. Originally, IQA metrics were mostly based on either physical \cite{PSNR, Angularerror} or perceptual \cite{NIQE,BRISQUE,SSIM,VIF,deltaE,mantiuk2011hdr} priors, but recently these priors are usually learned by using deep learning approaches from some type of observer data (e.g., mean opinion scores)~\cite{FLIP,LPIPS,PIEAPP,HyperIQA}. 


A parallel line of research to the development of IQA metrics is the study of their applicability to specific scenarios. In colour constancy \cite{vazquezcorral2009}, authors found that observers did not prefer the image with the smallest angular error against a white reference, but instead the one with a slightly bluish tint---the so-called blue bias \cite{Hurlbert_bluebias}. Similarly, \cite{Zamir21} showed that no IQA metric correlated well with the observers' choice in colour gamut mapping.

\myparagraph{Lighting perception.}
\label{subsec:related_work_human_perception_lighting}
Humans were shown to perceive lighting direction accurately on both synthetic and real stimuli \cite{koenderink2004light, pont2007matching}.  \cite{pont2007matching} demonstrated that human accuracy in lighting direction is correlated to the degree of collimation of the light source, explaining humans' high sensitivity to lighting position outdoors.  
However, humans have more difficulty detecting lighting inconsistencies as the scene complexity increases \cite{lopez2010measuring, tan2015perception, ramanarayanan2007visual}.  
%
According to \cite{boyaci2006cues, o2010influence, te2017perception}, observers mainly use cast shadows and highlights to determine the direction and intensity of light sources.  \cite{o2010influence} confirms that humans use geometry to estimate the lighting direction and that a globally convex shape increases their accuracy.  
%
\cite{anderson_image_2009, marlow_perception_2012} reveal that the perception of gloss is modulated by the complex interaction between the object geometry and the light field, affecting multiple parameters of specular reflections (size, contrast, sharpness, and depth).  
We use these works as inspiration to design our tasks and stimuli to bridge the gap between these human perception insights and IQA metrics commonly used to measure visual resemblance. 

%% file: sec/figs_latex/tex_fig_exp_tasks.tex
\begin{figure*}
   \centering
   \scriptsize
   \newlength{\mywidth}
   \setlength{\mywidth}{0.16\linewidth}
   \setlength{\tabcolsep}{3pt}
   \begin{tabular}{ p{0.55\textwidth} p{0.42\textwidth} }
    Task 1: Three renders of \{matte, glossy\} spheres are presented. Select the one that resembles the most the centre one by pressing the left or right arrow on the keyboard. \vspace{0.5em} & 
    Task 2: Two images of \{matte, glossy\} spheres are presented, select the one that seems the most realistic by pressing on the left or right arrow on the keyboard. \\
    \end{tabular}
    \begin{tabular}{ c c }
        \begin{tabular}{ rc c c }
            \rotatebox{90}{\hspace{0.25em}\hspace{2.7em}Task 1, diffuse} &
            \includegraphics[width=\mywidth]{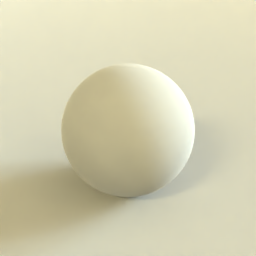} &
            \includegraphics[width=\mywidth]{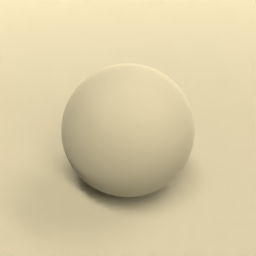} & 
            \includegraphics[width=\mywidth]{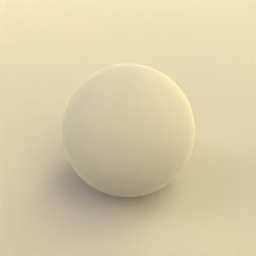} \\
        \end{tabular}
        \hspace{3em}
    &
        \begin{tabular}{ rc c }
            \rotatebox{90}{\hspace{0.25em}\hspace{2.7em}Task 2, diffuse} &
            \includegraphics[width=\mywidth]{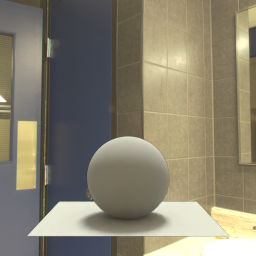} &
            \includegraphics[width=\mywidth]{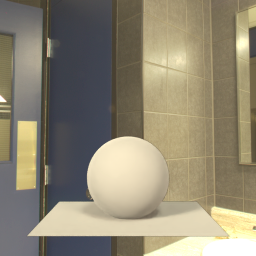}  \\
        \end{tabular}
    \\*[5em]
        \begin{tabular}{ rc c c }
            \rotatebox{90}{\hspace{0.25em}\hspace{2.7em}Task 1, glossy} &
            \includegraphics[width=\mywidth]{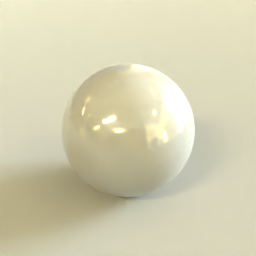} &
            \includegraphics[width=\mywidth]{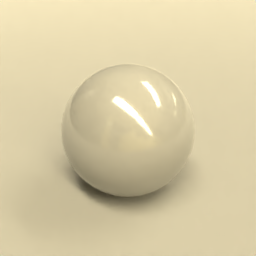} & 
            \includegraphics[width=\mywidth]{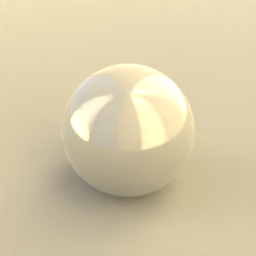} \\
        \end{tabular}
        \hspace{3em}
    & 
        \begin{tabular}{ rc c }
            \rotatebox{90}{\hspace{0.25em}\hspace{2.7em}Task 2, glossy} &
            \includegraphics[width=\mywidth]{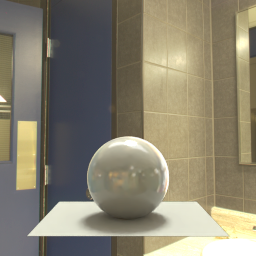} &
            \includegraphics[width=\mywidth]{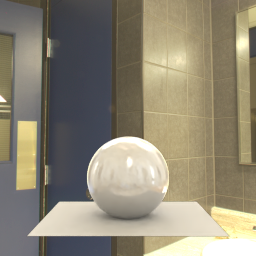}  \\
        \end{tabular} \\
    \end{tabular}
    \caption{Example of the accuracy (task 1; left) and plausibility (task 2; right) tasks, for the diffuse (top) and glossy (bottom) material, assigned to the observers during the experiment. The question asked to the observers is written above each example.} 
    \label{fig:exp_tasks}
\end{figure*}

%% file: sec/3_psychophys_exp.tex
\section{Psychophysical experiment}
\label{sec:psychophys_exp}

\input{sec/figs_latex/tex_fig_stimuli_individuel_exemple}
In this section, we briefly describe our psychophysical experiment, which was approved by the U.~Laval institutional Ethics Review Board, file \#2023-308.

\subsection{Tasks}
\label{subsec:psychophys_exp_stimuli}

We devise two tasks to evaluate how humans perceive lighting accuracy and plausibility, as illustrated in \cref{fig:exp_tasks}. 

\myparagraph{Task 1.} \Cref{fig:exp_tasks} (left) presents each observer with three images of a sphere on a ground plane: one lit with reference ground truth lighting (middle) and two with estimated lighting (left and right). The observer is then asked to choose which of the two renders resembles the reference the most. This task studies how much observers are able to \emph{match lighting to a reference}---in a sense, this is similar to what we ask IQA metrics to do. This will enable us, later on in \cref{sec:correlation}, to evaluate whether existing metrics perform similarly to humans \emph{at the same task}. 



\myparagraph{Task 2.} \Cref{fig:exp_tasks} (right) presents each observer with two images of a virtual scene (sphere on a plane) embedded within the background image from which lighting was estimated. The observer is then asked to select the most realistic of the two images. As opposed to task 1, no render with ground truth lighting is shown for this task. This task studies how observers \emph{judge the virtual objects in context}, without having to rely on a reference. 


To better distinguish the factors impacting human perception of lighting, we formulate four variants for each of those two tasks. 
The first two variants separate indoor and outdoor scenes, as both cases are generally addressed differently in the literature. The last two variants use different materials applied to the virtual object: one diffuse and one glossy. The goal is to bring the observer to focus on the low and high frequencies aspects of lighting, respectively. 

\subsection{Stimuli}
\label{sec:stimuli}

\myparagraph{Virtual objects.}
\label{subsubsec:psychophys_exp_stimuli_synthetic_scene}
%
To avoid potential issues with compositing errors, mismatching semantics and/or geometry, social biases of observers, etc., the virtual objects rendered are a simple sphere lying on a plane~\cite{o2010influence}. The sphere is seen closely (task 1, \cref{fig:exp_tasks} left) and farther away (task 2, \cref{fig:exp_tasks} right). The sphere is positioned farther away in task 2 to allow the observers to see the background, as the sphere and plane fill the render in task 1. The Disney Principled BRDF~\cite{Burley2012PhysicallyBasedSA} is used for both diffuse (roughness \num{1.0} and specularity \num{0.0}) and glossy (roughness \num{0.1} and specularity \num{1.0}) materials, with an albedo of \num{0.18} in both cases.
The images are generated with the Cycles physically-based rendering engine, and then reexposed and tonemapped with $\gamma = 2.4$ to be displayed on the monitor. 

\myparagraph{Lighting estimation methods.}
\label{subsubsec:psychophys_exp_stimuli_lighting_estimation_methods}
We include recent learning-based lighting estimation methods for both the in- and outdoor scenarios, including the current state-of-the-art by Weber~\etal~\cite{weber2022editable} (indoor), competitive GAN-based methods EverLight \cite{Dastjerdi_2023_ICCV} (indoor and outdoor) and StyleLight \cite{wang2022stylelight} (indoor), and approaches predicting parametric lights: Gardner~\etal~\cite{gardner2019deep} (indoor) and Zhang~\etal~\cite{zhang2019all} (outdoor). We also include Khan~\etal~\cite{khan2006image} as an example of a simpler, non-learning technique, since it estimates the lighting by projecting the input image onto a sphere and then mirroring it to generate an environment map. See \cref{fig:stimuli_individuel} for examples of lighting estimated by the selected methods.  

\myparagraph{Input scenes.}
\label{subsubsec:psychophys_exp_stimuli_scenes}
To ensure the diversity of the selected scenes, we compute the first-order spherical harmonics coefficients and cluster them using $k$-means, with $k=25$. We select the \valeur{\num{25}} panoramas closest to each cluster centre. Finally, we extract a perspective image of limited field of view from each HDR panorama in our datasets by simulating a camera taking a picture within the environment.
\subsection{Experimental setup}
\label{subsubsec:psychophys_exp_experimental_setup_material}

\myparagraph{Hardware.}
The experiment was conducted in a controlled lab setting to ensure the uniformity of the data collected. It was carried out in a matte black room with a standard keyboard placed on a desk. The monitor was set to sRGB and was the only light source in the room.  
The observers were seated $\approx$\valeur{\SI{70}{\cm}} away from the monitor, which results in a \valeur{\ang{11.5}}/\valeur{\ang{17}} visual angle per image for task 1/2 respectively.


\myparagraph{Procedure.}
\label{subsubsec:psychophys_exp_experimental_setup_procedure}
Two distinct independent two-alternative forced choice (2AFC) tasks are given to the observers sequentially.  The pairs of all combinations of the stimuli produced by the lighting estimation methods (\valeur{\num{5}} indoor and \valeur{\num{3}} outdoor) for the \valeur{\num{25}} different scenes are presented to the observers. Thus, a total of \valeur{\num{250}} images (\valeur{\num{10}} method combinations $\times$ \valeur{\num{25}} scenes) for the indoor case and  \valeur{\num{75}} images (\valeur{\num{3}} method combinations $\times$ \valeur{\num{25}} scenes) for the outdoor case are used.  The order and the placement (left or right) of each method are randomised to reduce potential biases.


\myparagraph{Participants.}
\label{subsubsec:psychophys_exp_experimental_setup_participants}
A total of \valeur{\num{49}} unique observers (\valeur{\num{33}}M/\valeur{\num{16}}F, ages ranging from \valeur{\numrange[range-phrase = --]{24}{63}}) with normal or corrected-to-normal vision participated in the study. All the observers were tested for colour blindness using the Ishihara test.
Not all observers participated in all experiments, resulting in indoor experiments performed by \num{30}/\num{31} observers for the diffuse/glossy versions respectively and outdoor experiments by \num{12} observers. Participants were allowed to do each experiment at most once.
None of the authors took part in the study. Each experiment takes \valeur{\SI{\sim 25}{\minute}} for the indoor and \valeur{\SI{\sim 5}{\minute}} for the outdoor cases. No time restriction is imposed to avoid inducing stress and bias.


%% file: sec/figs_latex/tex_fig_stimuli_individuel_exemple.tex
\begin{figure*}
\scriptsize
\setlength{\tabcolsep}{0.5pt}
\newlength{\tmplength}
\setlength{\tmplength}{0.095\linewidth}
\begin{tabular}{lccccccccccc}
     &                                                                                                 
     GT indoor &      
     Weber~\cite{weber2022editable} &
     EverLight~\cite{Dastjerdi_2023_ICCV} & 
     StyleLight~\cite{wang2022stylelight} &  
     Gardner~\cite{gardner2019deep} &                 
     Khan~\cite{khan2006image} &
     GT outdoor & 
     EverLight~\cite{Dastjerdi_2023_ICCV} & 
     Zhang~\cite{zhang2019all} & 
     Khan~\cite{khan2006image} \\
\rotatebox{90}{\hspace{0.25em}Task 1, diffuse} & 
\includegraphics[width=\tmplength]{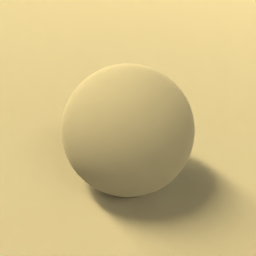} & 
\includegraphics[width=\tmplength]{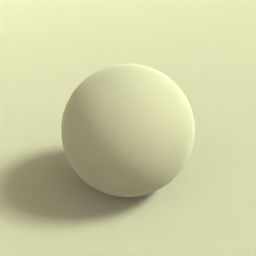} & 
\includegraphics[width=\tmplength]{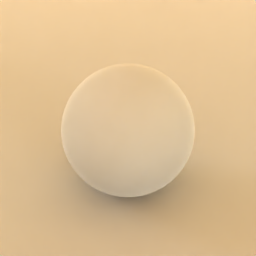} & 
\includegraphics[width=\tmplength]{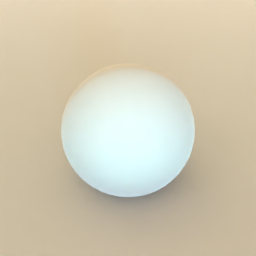} & 
\includegraphics[width=\tmplength]{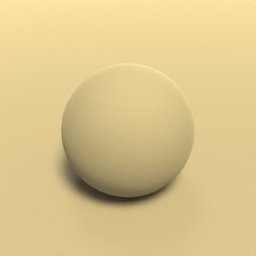} & 
\includegraphics[width=\tmplength]{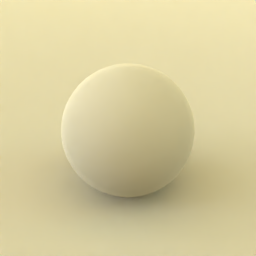} & 
\includegraphics[width=\tmplength]{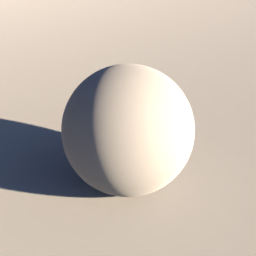} & 
\includegraphics[width=\tmplength]{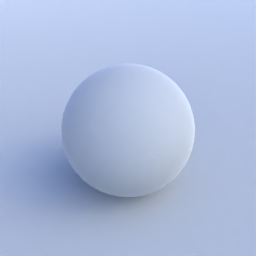} & 
\includegraphics[width=\tmplength]{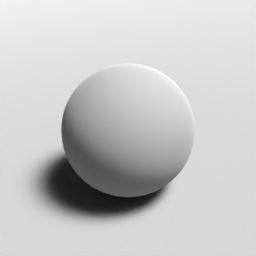} & 
\includegraphics[width=\tmplength]{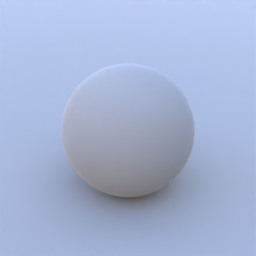} \\
\rotatebox{90}{\hspace{0.35em}Task 1, glossy} &
\includegraphics[width=\tmplength]{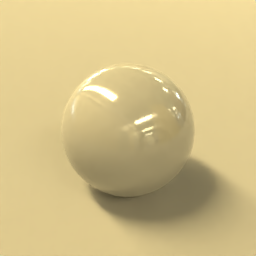} &  
\includegraphics[width=\tmplength]{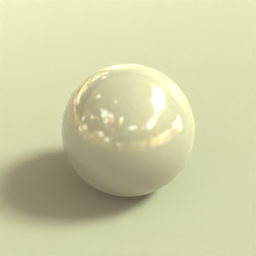} &  
\includegraphics[width=\tmplength]{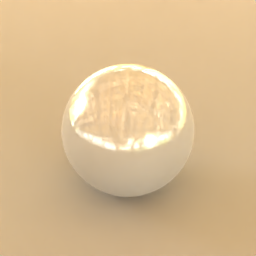} &  
\includegraphics[width=\tmplength]{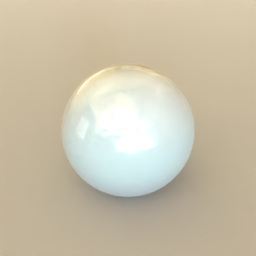} &  
\includegraphics[width=\tmplength]{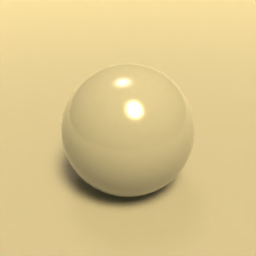} &  
\includegraphics[width=\tmplength]{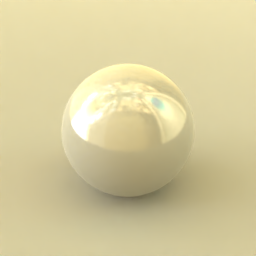} &  
\includegraphics[width=\tmplength]{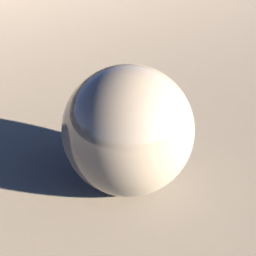} &  
\includegraphics[width=\tmplength]{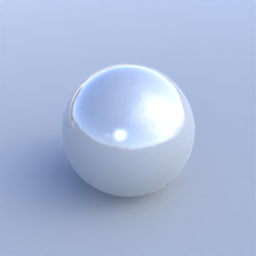} &  
\includegraphics[width=\tmplength]{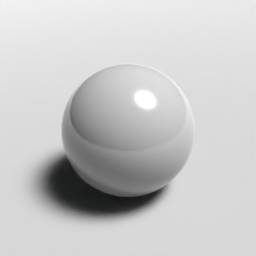} &  
\includegraphics[width=\tmplength]{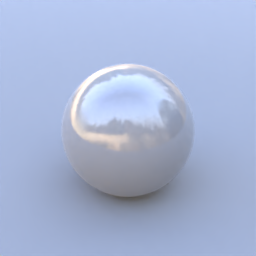} \\
\rotatebox{90}{\hspace{0.25em}Task 2, diffuse} &
\includegraphics[width=\tmplength]{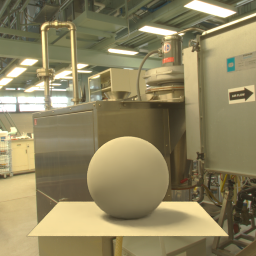} &    
\includegraphics[width=\tmplength]{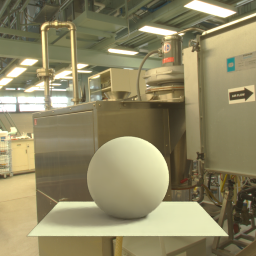} &    
\includegraphics[width=\tmplength]{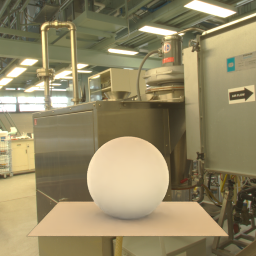} &    
\includegraphics[width=\tmplength]{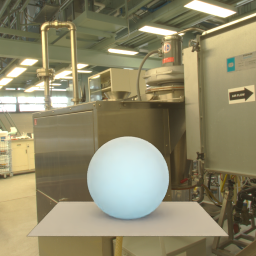} &    
\includegraphics[width=\tmplength]{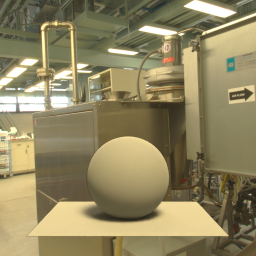} &    
\includegraphics[width=\tmplength]{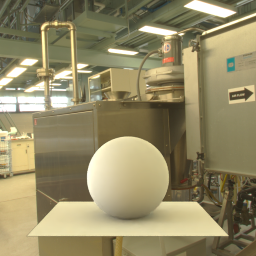} &    
\includegraphics[width=\tmplength]{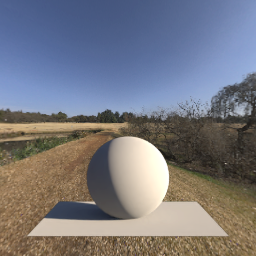} &    
\includegraphics[width=\tmplength]{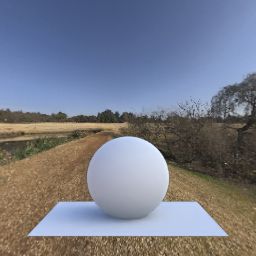} &    
\includegraphics[width=\tmplength]{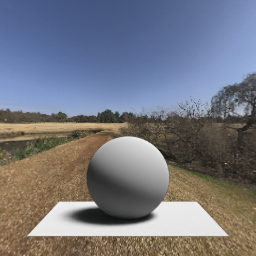} &    
\includegraphics[width=\tmplength]{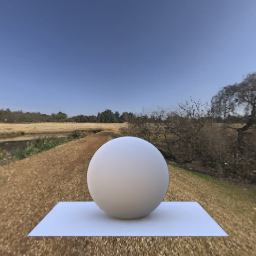} \\
\rotatebox{90}{\hspace{0.35em}Task 2, glossy} &
\includegraphics[width=\tmplength]{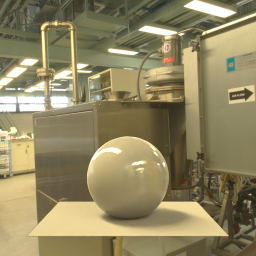} &     
\includegraphics[width=\tmplength]{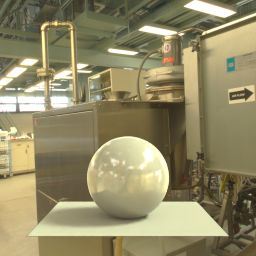} &     
\includegraphics[width=\tmplength]{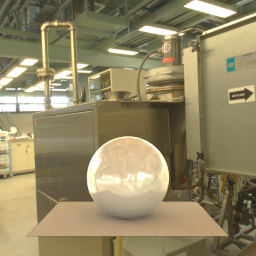} &     
\includegraphics[width=\tmplength]{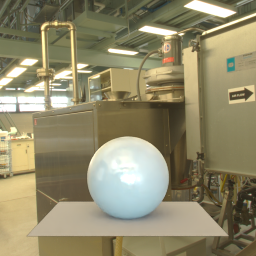} &     
\includegraphics[width=\tmplength]{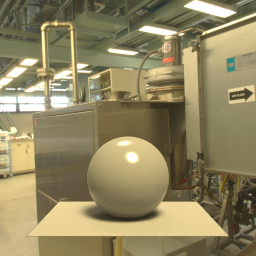} &     
\includegraphics[width=\tmplength]{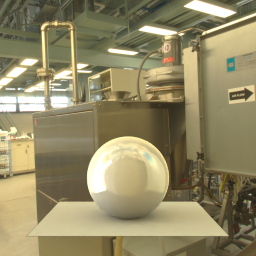} &     
\includegraphics[width=\tmplength]{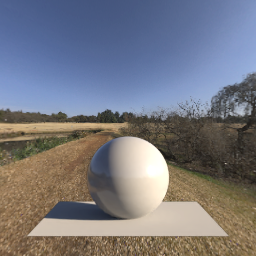} &     
\includegraphics[width=\tmplength]{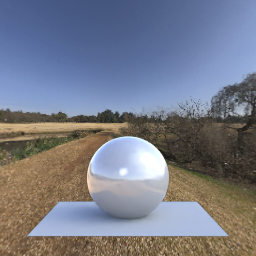} &     
\includegraphics[width=\tmplength]{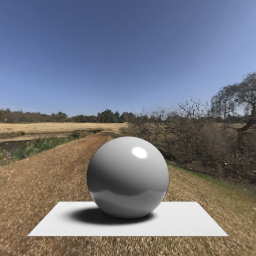} &     
\includegraphics[width=\tmplength]{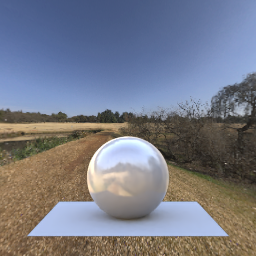} \\
\rotatebox{90}{Envmap} & 
\includegraphics[width=\tmplength,trim=10px 10px 10px 10px,clip]{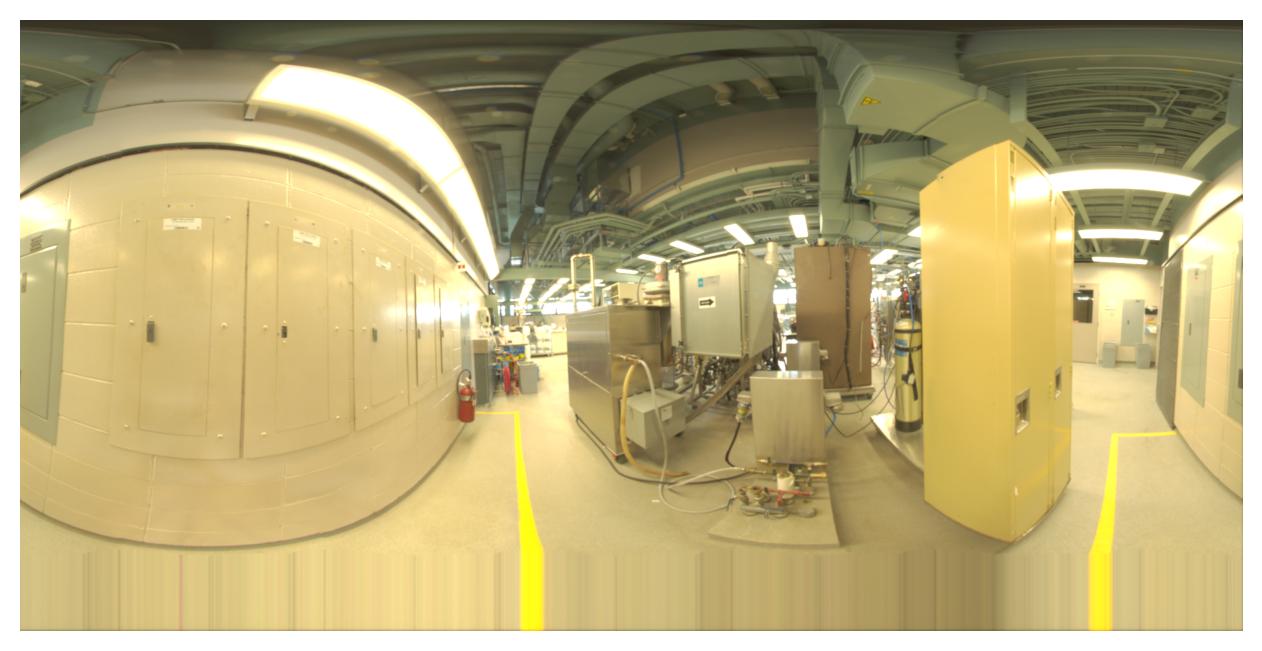} &                
\includegraphics[width=\tmplength,trim=10px 10px 10px 10px,clip]{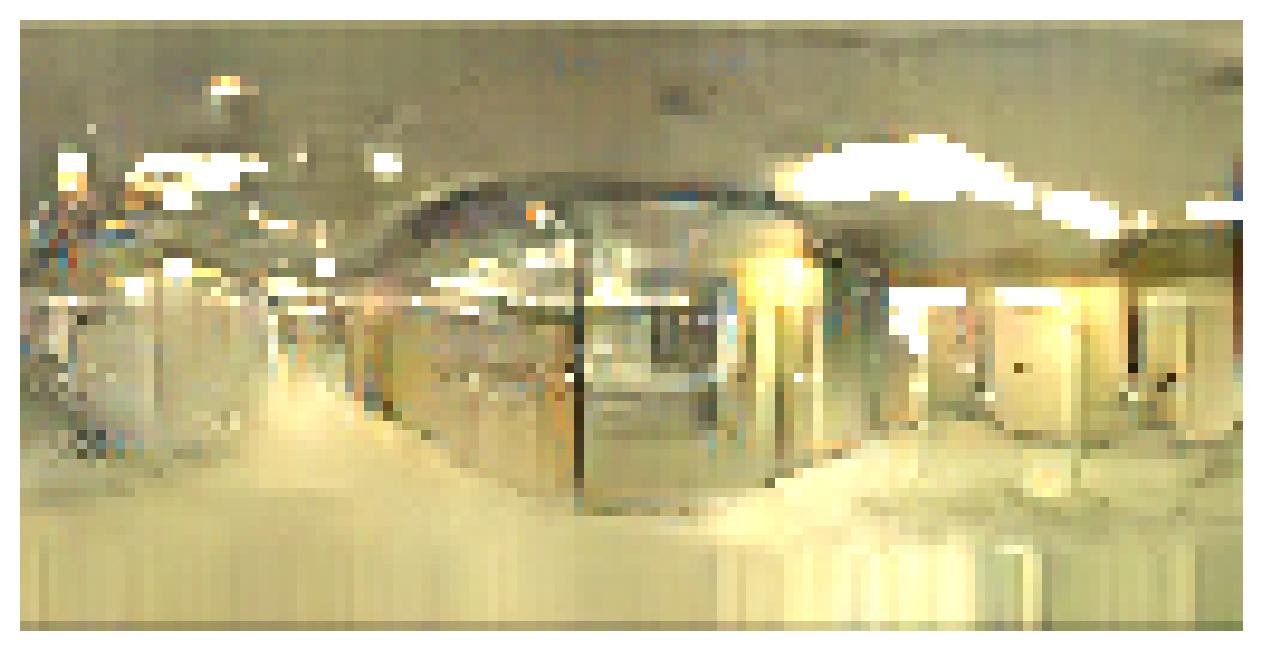} &                
\includegraphics[width=\tmplength,trim=10px 10px 10px 10px,clip]{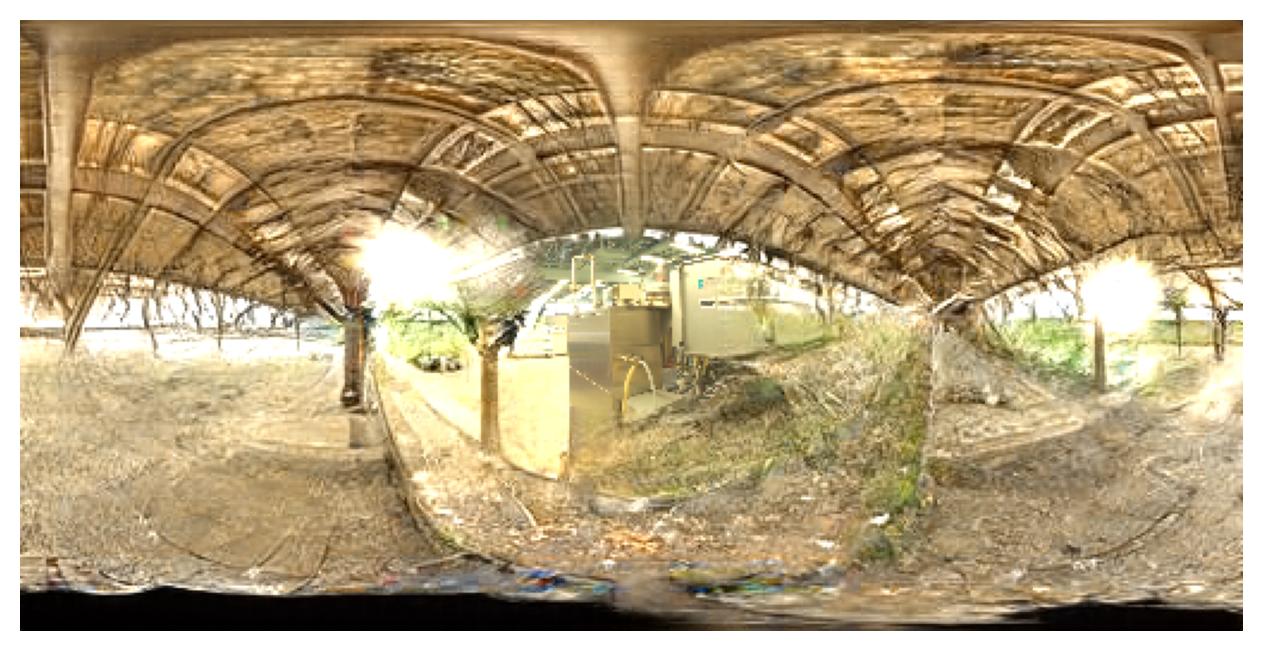} &                
\includegraphics[width=\tmplength,trim=10px 10px 10px 10px,clip]{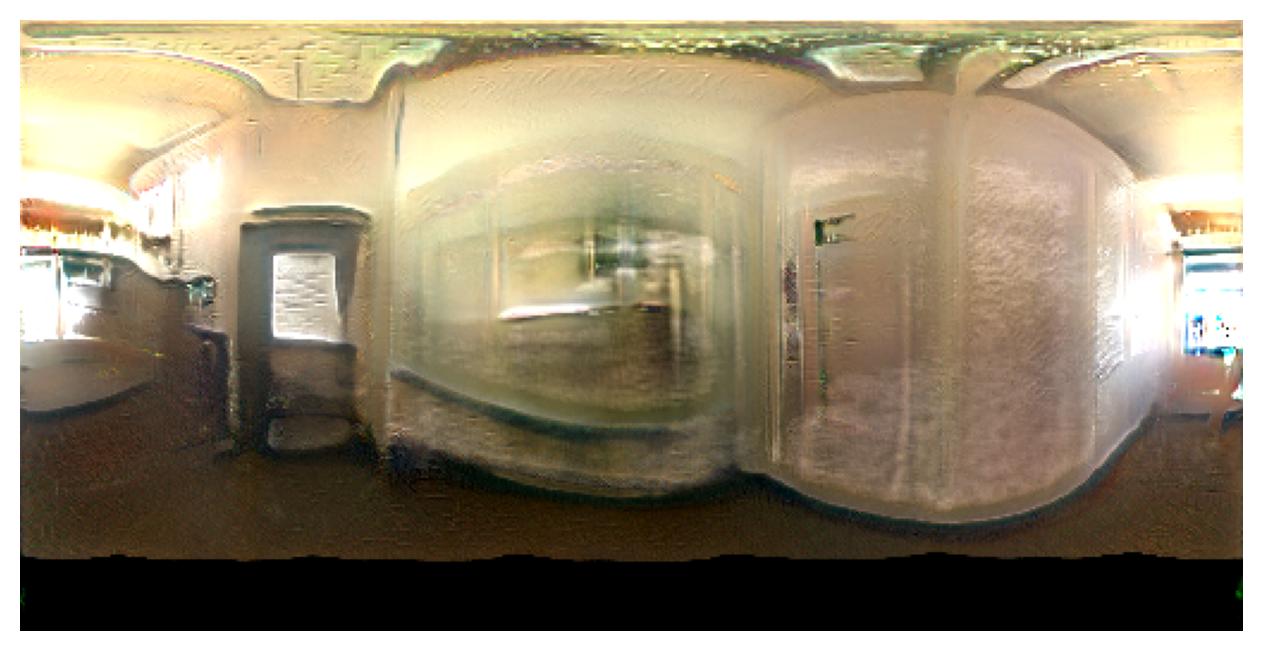} &                
\includegraphics[width=\tmplength,trim=10px 10px 10px 10px,clip]{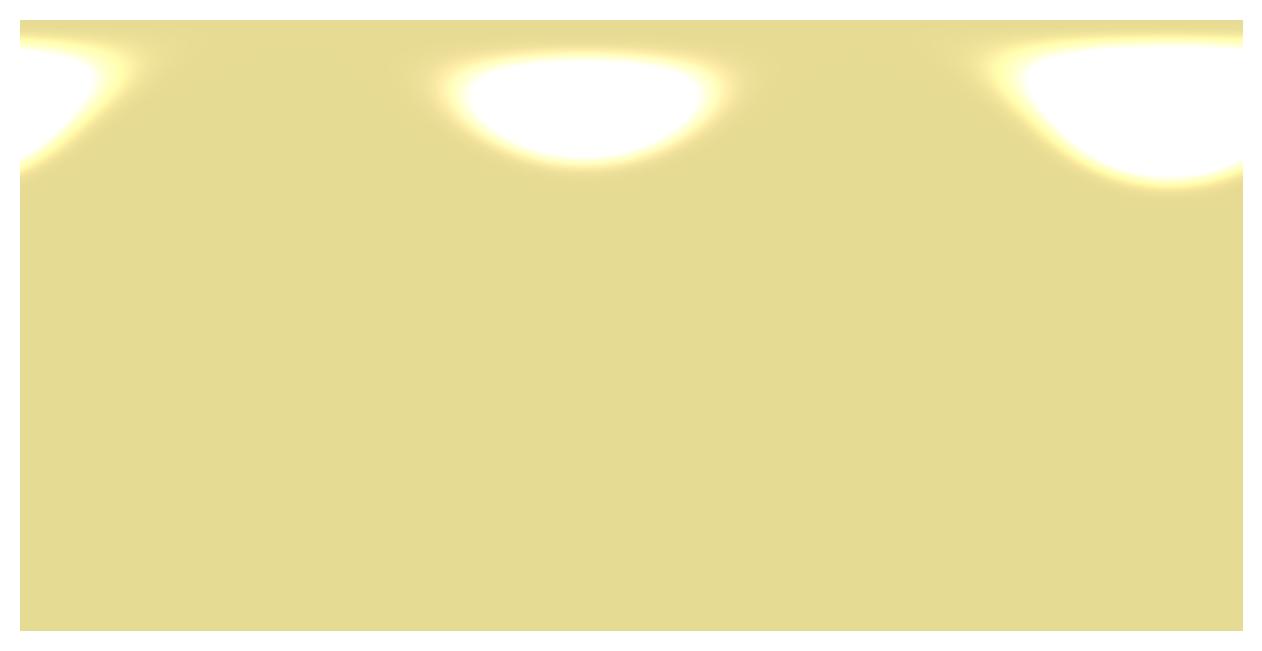} &                
\includegraphics[width=\tmplength,trim=10px 10px 10px 10px,clip]{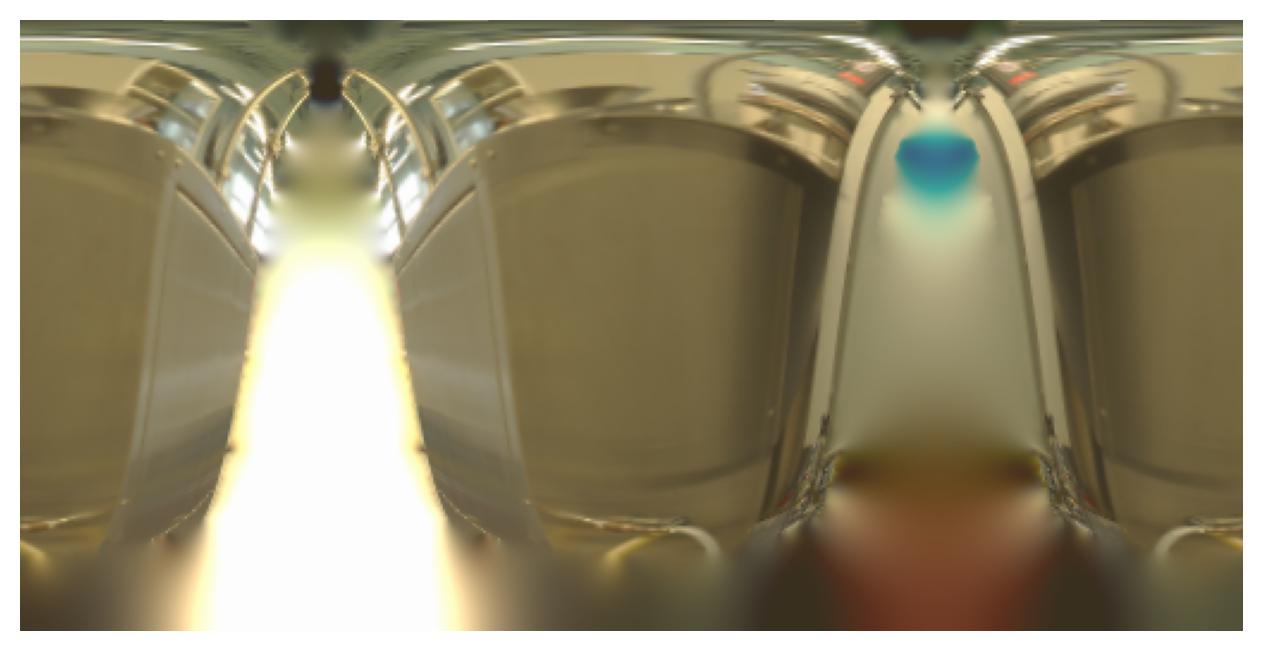} &                
\includegraphics[width=\tmplength,trim=10px 10px 10px 10px,clip]{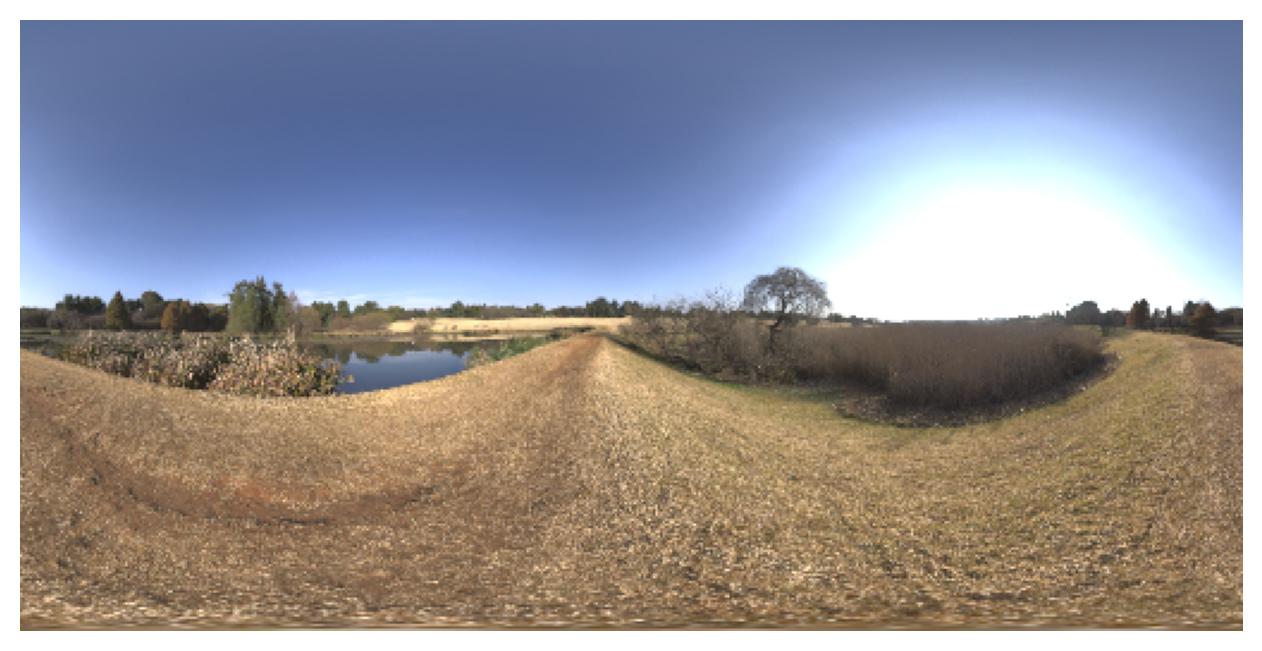} &                
\includegraphics[width=\tmplength,trim=10px 10px 10px 10px,clip]{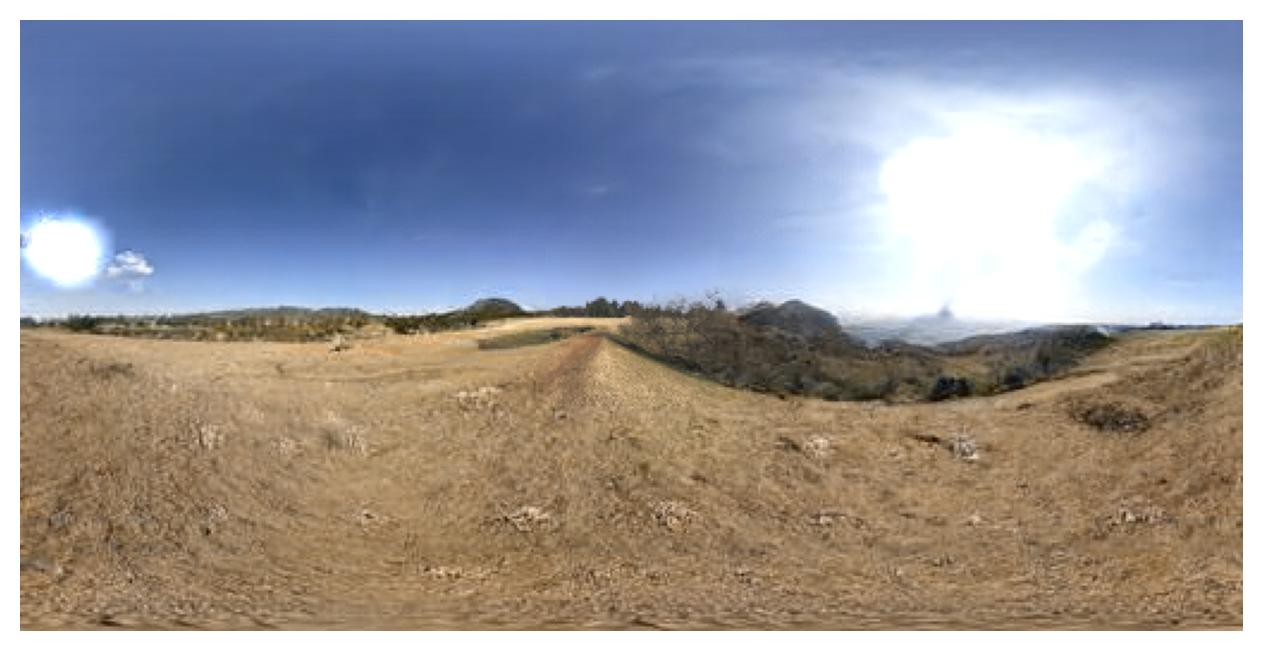} &                
\includegraphics[width=\tmplength,trim=10px 10px 10px 10px,clip]{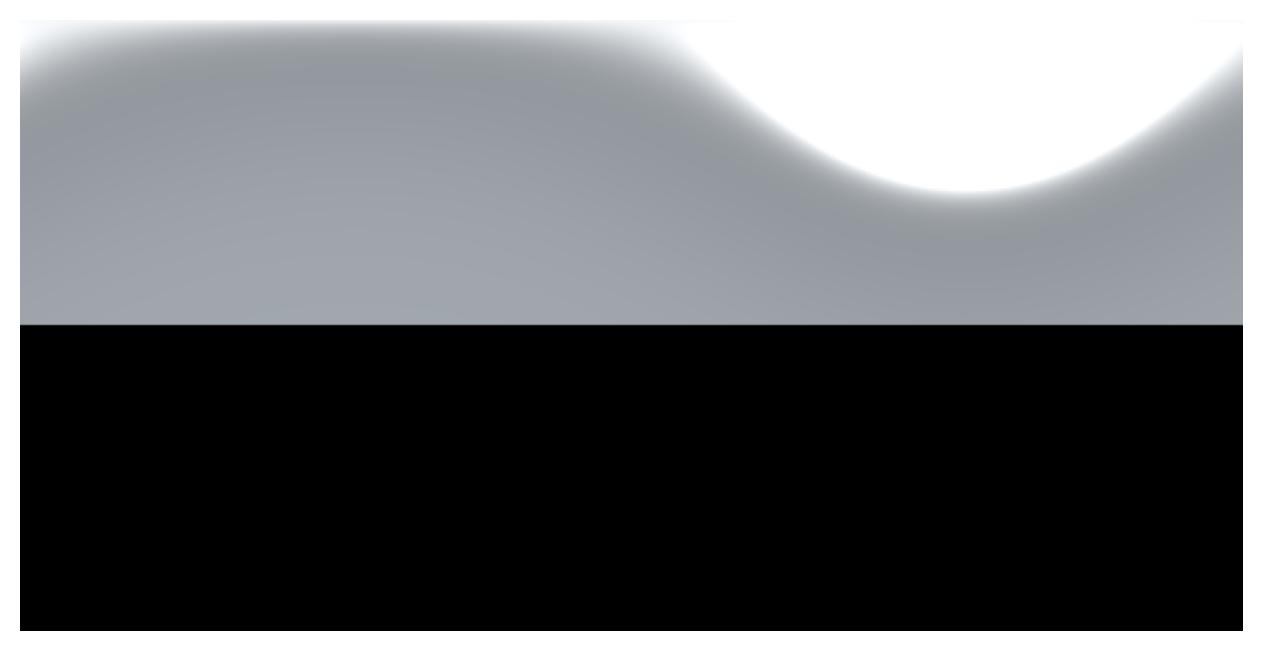} &                
\includegraphics[width=\tmplength,trim=10px 10px 10px 10px,clip]{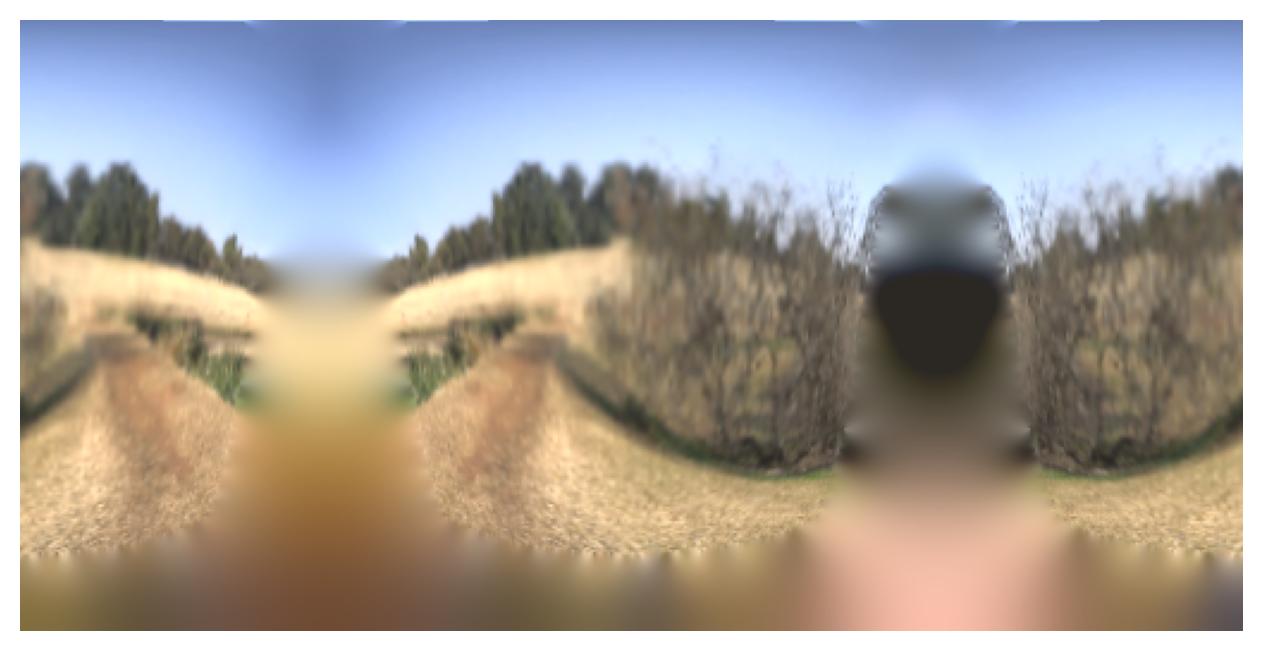} \\
\end{tabular}
\caption{Example of the stimuli produced for a scene by each of the lighting estimation methods (columns), for the different tasks and experiments (rows).  The last row corresponds to the estimated lighting (projected to an equirectangular format) by each of the methods and used for the renders.  The ``GT'' columns correspond to the ground truth---note that it is not used in task 2 and shown here only for reference.}
\label{fig:stimuli_individuel}
\end{figure*}

%% file: sec/4_results.tex
\begin{figure*}[t]
  \centering
  \includegraphics[width=0.95\linewidth]{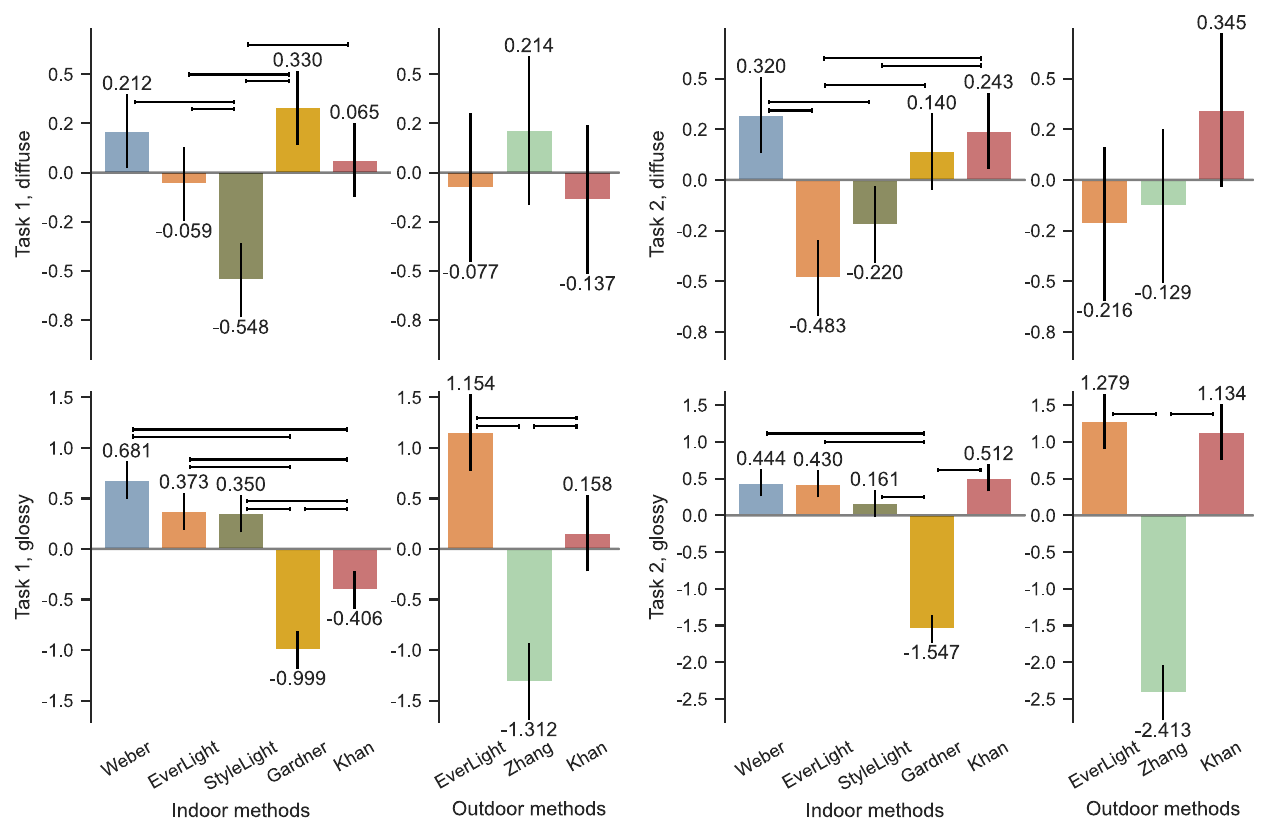}
  \vspace{-0.8em}
  \caption{Thurstone Case V Law of Comparative Judgement ($z$-scores) for all the observers as a function of the different lighting estimation methods (bars), for the different materials (rows) and tasks (columns).  A positive score indicates that observers generally prefer the stimuli rendered with the lighting estimation method, and not preferred when the score is negative.  The scores of all the methods for an experiment sum to \num{0}.  The brackets above indicate pairs of methods for which the perceptual difference is statistically significant.  Error bars correspond to \SI{95}{\%} confidence interval. }
  \label{fig:psychophysical_experiment_in_and_out_results_all_observers_per_method}
\end{figure*}

\section{Evaluating lighting estimation methods with perceptual data}
\label{sec:results}

In this section, we evaluate state-of-the-art lighting estimation methods in light of our perceptual data acquired in \cref{sec:psychophys_exp}. We begin by describing our statistical analysis approach, then highlight several key observations in the subsequent analysis. 


\subsection{Data processing}
\label{subsec:results_data_processing}

Since observers were asked to select an image produced by a particular lighting estimation method, we assign, for each method, a \num{1} if its result was selected and a \num{0} otherwise. For each of the individual experiments, the resulting $z$-scores for the lighting estimation methods are computed according to the Thurstone Case V Law of Comparative Judgement model~\cite{thurstone2017} and are shown in \cref{fig:psychophysical_experiment_in_and_out_results_all_observers_per_method}. 
This method receives as input a matrix $\vb{C} \in \mathbb{R}^{N \times N}$ of paired comparison data, where element $\vb{C}_{i,j}$ indicates how many times method $i$ was preferred over method $j$, and outputs a scaling $z$-score for all $N$ methods compared. This $z$-score is defined such that a higher value indicates higher observer preference, with a sum over all methods of 0 (so methods that are consistently not preferred get a negative value). We compute the $z$-score individually for all images from a scene (set of stimuli images produced by all $N$ lighting estimation methods with the same scene given as input). The mean across all scenes is then taken to describe the performance of each method in a specific experiment. 

The uncertainties, illustrated as error bars in \cref{fig:psychophysical_experiment_in_and_out_results_all_observers_per_method}, correspond to the \SI{95}{\%} confidence interval using Montag's method \cite{Montag2006}. 
These confirm that the number of participants is sufficient, as statistical significance is reached (indicated by the horizontal brackets) for most of the experiments. However, the uncertainty is higher for the outdoor experiments, as it is only proportional to the number of observers.
We also compute Fleiss' $\kappa$~\cite{fleiss1971measuring} and Kuder-Richardson-20 (KR20)~\cite{kuder1937theory} between the data of each observers to demonstrate the agreement and consistency, respectively, in the supplementary material, which confirm that the number of participants for the outdoor experiment is sufficient.

\subsection{Observations}
\label{subsec:results_user_study_analysis}


At first glance, \cref{fig:psychophysical_experiment_in_and_out_results_all_observers_per_method} reveals that the observers' judgement varies across tasks and materials. This suggests that judging the accuracy (task 1) vs plausibility (task 2) of lighting are cognitively different tasks. Therefore, humans likely focus on different cues when looking at images. We now highlight several key observations emanating from \cref{fig:psychophysical_experiment_in_and_out_results_all_observers_per_method}. See the supplemental for more lighting estimation methods rankings. 


\myparagraph{Solely predicting light sources in parametric format is insufficient.}
First, we observe that methods focusing exclusively on predicting light sources in parametric format (Gardner~\etal~\cite{gardner2019deep} indoors and Zhang~\etal~\cite{zhang2019all} outdoors) do perform well when matching diffuse renders to ground truth (task 1, diffuse), but are otherwise disliked by observers when the scene contains glossy objects. 


\myparagraph{Methods should predict a combination of HDR lighting and plausible textures.}
Overall, Weber~\etal~\cite{weber2022editable} seems to be the preferred method across all indoor experiments. 
It combines HDR lighting estimation that is similarly accurate to that of Gardner~\etal~\cite{gardner2019deep} (task 1, diffuse) with plausible textures closely matching the ground truth (task 1, glossy). Results from task 2 also reveal that its composites are consistently judged to be plausibly blended with their surroundings. Both EverLight~\cite{Dastjerdi_2023_ICCV} and StyleLight~\cite{wang2022stylelight} are GAN-based approaches that aim at predicting HDR lighting and textures as well. Whilst they do succeed at predicting visually pleasing textures that create believable reflections on glossy materials, their drop in performance when diffuse materials are used indicates that they seem to struggle at generating accurate (task 1, diffuse) and plausible (task 2, diffuse) HDR lighting.
These results indicate that accurate HDR lighting estimation is important in the diffuse spheres experiments, and that plausible textures seem more important for glossy objects.

\myparagraph{Matching ground truth lighting matters less when judging the plausibility of composites.}
%
The older method of Khan~\etal~\cite{khan2006image} is generally expected to perform less well than more recent methods since it creates distorted, implausible LDR textures.  
Indeed, results from task 1 show that its lighting estimates do not perceptually match the ground truth, especially in the case of glossy materials. 
However, when considering the plausibility of composited objects in the absence of a reference (task 2), the trend is reversed: the simple method of \cite{khan2006image} yields results on par with those of \cite{weber2022editable} (indoors) and \cite{Dastjerdi_2023_ICCV} (outdoors, glossy). Surprisingly, it even outperforms all the recent methods in one scenario (outdoors, diffuse)! This suggests that, even if the predicted lighting is perceived to be inaccurate (task 1), the accuracy is not a priority for the observers when judging the realism of inserted virtual objects.

%% file: sec/5_measure_comparison.tex
\section{Measuring the agreement between IQA metrics and perceptual data}
\label{sec:correlation}

\begin{figure*}[t]
  \centering
   \includegraphics[width=1.0\linewidth]{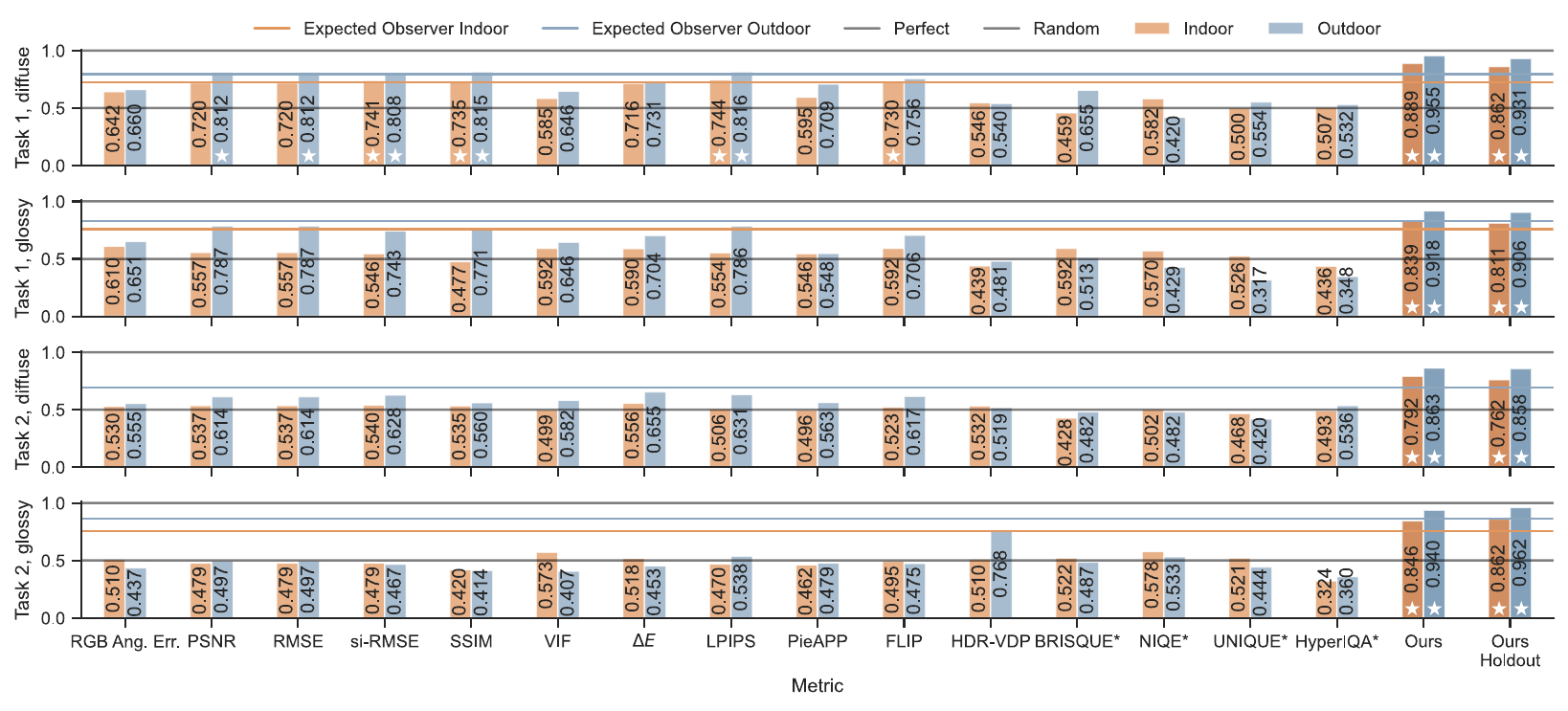}
   \vspace{-2em}
   \caption{Agreement between the observer scores and the metric scores (columns) for all the lighting estimation methods (indoor: \valeur{orange} bars; outdoor: \valeur{blue} bars), for the different types of experiments (rows).  The lower horizontal \valeur{grey} line is set at chance level (\valeur{\num{\sim 0.5}}) and the higher one corresponds to the perfect observer (set at \valeur{\num{1.0}}).  The \valeur{orange} (indoor) and \valeur{blue} (outdoor) lines correspond to the expected observer agreement score (for task 2, diffuse, blue and orange lines are overlapped).   The stars indicate methods that have an agreement score equal or superior to the expected observer. ``Ours'' and ``Ours Holdout'' refer to our learned metric combination, see \cref{sec:metric} for more details.  The No-Reference IQA metrics are indicated by asterisks.}
   \label{fig:correlation_in_and_out_observers_metrics}
\end{figure*}

In this section, we pit existing metrics against our perceptual data to determine the degree to which they agree about lighting perception. In particular, we consider Full-Reference metrics (RGB angular error~\cite{Angularerror}, PSNR~\cite{PSNR}, RMSE, si-RMSE, SSIM~\cite{SSIM}, VIF~\cite{VIF}, $\Delta{}E$~\cite{deltaE}, LPIPS~\cite{LPIPS}, PieAPP~\cite{PIEAPP}, FLIP~\cite{FLIP}, HDR-VDP3~\cite{mantiuk2023hdr}) as they are the standard in the community for lighting estimation method evaluation.
We also include No-Reference metrics (BRISQUE~\cite{BRISQUE}, NIQE~\cite{NIQE}, UNIQUE~\cite{zhang2021uncertainty}, HyperIQA~\cite{HyperIQA}) to evaluate whether they better match human perception. All metrics are computed on the tonemapped stimuli (as seen by the observers, c.f. \cref{sec:stimuli}) except HDR-VDP3 which uses HDR renders (before gamma correction).



\subsection{Agreement score}
\label{subsec:results_correlation_data_processing}

We aim to study if our collected perceptual data agrees with the results given by the metrics. To do so, we define an agreement score between a metric (or observer) and a hypothetical \emph{perfect observer}. Larger values of this agreement score will mean a higher agreement with the perfect observer.

Let $\varphi_{a, b}^{(i)}$ be the binary choice of observer $i \in \{1, \ldots, n\}$ between stimuli $a$ and $b$, representing the output of two different lighting estimation methods for the same scene:
\begin{align}
    \varphi_{a, b}^{(i)} \overset{\Delta}{=} 
    \begin{cases} 
    1 & \text{if $a$ was selected over $b$},\\  
    0 & \text{otherwise}.
    \end{cases}
    \label{eq:input_data}
\end{align}
Note that this relation is symmetric, i.e., $\varphi_{b, a}=1-\varphi_{a, b}$ but we slightly abuse notation and write $\varphi_{a, b}$ for both cases.
The proportion at which all $n$ observers select each stimulus is computed as the mean $\overline{\varphi}_{a, b} = \frac{1}{n} \sum_{i=1}^n \varphi^{(i)}_{a,b}$.
We are interested in the comparisons where most of the observers selected $a$, so all values of $\overline{\varphi}_{a, b} < 0.5$ (i.e. when $b$ is the most selected) are set to $0$. The agreement score $\omega^{(i)}$ between observer $i$ and the mean choice of all observers $\overline{\varphi}_{a, b}$ is
\begin{align}
    \omega^{(i)} = \frac{\sum_{(a, b)} \overline{\varphi}_{a, b} \, \varphi_{a, b}^{(i)}}{\sum_{(a, b)} \overline{\varphi}_{a, b}} \,.
    \label{eq:agreement_score}
\end{align}
%
Finally, the \emph{expected observer agreement} $\overline{\omega} = \frac{1}{n} \sum_{i=1}^{n} \omega^{(i)}$ is the mean agreement over all observers. 

%


A hypothetical \emph{perfect observer} is defined as an observer who makes the exact same choices as the average observer. In this case, $\omega^{(i)} = 1$ and is the upper-bound score. Conversely, a \emph{random observer} is defined as an observer who would randomly select images. Since pairs of images are presented, its agreement score is $\omega^{(i)} = 0.5$. 
We removed all data from individual observers with an agreement score $\omega^{(i)} < 0.5$ in at least one experiment (\num{5} individuals in our study) since they likely have misunderstood the task. 

We also define the agreement score for each IQA metric. For this, we recompute \cref{eq:agreement_score} by replacing observer choices $\varphi$ with the ones made by metrics. Note that metrics choose the best image by selecting the image most (dis)similar to ground truth, even though observers do not always have access to it (e.g., task 2).


\subsection{Observations}
\label{subsec:results_correlation_analysis}

The agreement scores for each IQA metric are shown alongside the expected observer agreement in \cref{fig:correlation_in_and_out_observers_metrics} for all experiments.  
The \valeur{grey} lines correspond to the perfect (upper) and random (lower) observers. The \valeur{orange}/\valeur{blue} lines correspond to the expected observer agreement score for the indoor/outdoor experiments, respectively. A higher metric agreement score indicates that it tends to select the same stimuli as the observers.


Note that metrics generally select the same images irrespective of the task since the geometry, material and lighting in the renders are the same. 
However, perceptual results are task-dependent (c.f. \cref{subsec:results_user_study_analysis}), which suggests that humans and metrics use different processes when evaluating lighting in different images, shown by the difference in agreement with the metrics and humans in \cref{fig:correlation_in_and_out_observers_metrics}.

\myparagraph{Metrics tend to perform well on task 1 with diffuse materials.}
The stars on \cref{fig:correlation_in_and_out_observers_metrics} indicate the metrics with agreement scores equal or superior to the expected observer, indicating that they agree with human perception.  This occurs only with si-RMSE, SSIM, LPIPS, and FLIP for task 1 diffuse (indoor), and PSNR, RMSE, si-RMSE, SSIM, and LPIPS for task 1 diffuse (outdoor). Thus, these metrics may indeed be appropriate for these specific tasks, where observers are asked to match objects with diffuse appearance. 

\myparagraph{Metrics do not agree with human perception for lighting estimation for the rest of the tasks.}
\Cref{fig:correlation_in_and_out_observers_metrics} shows that IQA metrics do not usually agree with observers. In many cases, especially for task 1 glossy and task 2 diffuse and glossy, metrics even perform on par with a random observer. Our study demonstrates that we should not rely on such metrics, in the general case, for evaluating lighting estimation algorithms as they do not accurately reflect human judgement.  

\subsection{Correlation}
\label{subsec:results_correlation_stat_test}


\captionsetup[table]{name=Table}
\crefname{table}{tab.}{tabs.}
\Crefname{table}{Tab.}{Tabs.}
\setcounter{table}{0}

We computed Spearman's $\rho$ and Kendall's $\tau$ statistical tests to measure the correlation between the binary observers' choices and the estimated metric choices, obtained from fitting a logistic curve as recommended by \cite{wang2003multiscale}.  \Cref{tab:table_stat_test_best_metrics_results} shows the best two metrics for each indoor experiment.

\input{sec/supp_figs_latex/statistical_tests_best_metrics_results}

\captionsetup[table]{name=Figure}
\crefname{table}{fig.}{figs.}
\Crefname{table}{Fig.}{Figs.}
\setcounter{table}{\thefigure}
\setcounter{figure}{5} 

Our metric always obtains the highest correlation scores, matching the perception data best and corroborating the analysis from the agreement scores in \cref{fig:correlation_in_and_out_observers_metrics}. We note there is still room for improvement since scores are lower than \num{0.8}, which is the threshold typically considered as reliable by those tests, thus paving the way for future work.

Overall, no metric can accurately judge the lighting estimation in all contexts, either on accuracy (task 1) or plausibility (task 2), for all materials (diffuse or glossy), in all domains (indoor and outdoor). 

%% file: sec/supp_figs_latex/statistical_tests_best_metrics_results.tex
\begin{table}[t!]
\footnotesize
\centering
\caption{Correlation scores of our metric and the next best metric (excluding ``Ours Holdout``) for Spearman's $\rho$ and Kendall's $\tau$ tests, for all indoor experiments. }
\label{tab:table_stat_test_best_metrics_results}
\begin{tabular}{lllcc}
\toprule
Task & Material & Metric & Spearman's $\rho$ & Kendall's $\tau$ \\
\midrule
Task 1 & Diffuse & Ours &  \textbf{0.689} & \textbf{0.572} \\
       &        & LPIPS &  0.529 &  0.440 \\
       & Glossy & Ours &  \textbf{0.724} & \textbf{0.601} \\
       &        & RGB Ang. Err. &  0.226 &  0.187 \\
Task 2 & Diffuse & Ours &  \textbf{0.711} &  \textbf{0.593}\\
       &        & BRISQUE &  0.145 &   0.121\\
       & Glossy & Ours &  \textbf{0.753} & \textbf{0.628} \\
       &        & HyperIQA &  0.371 &  0.308 \\
\bottomrule
\end{tabular}
\vspace{-1em}
\end{table}


%% file: sec/6_metric.tex
\section{Learning a metric combination}
\label{sec:metric}


In this section, we propose to learn task-specific functions which map existing IQA metrics to perceptual data. This is similar to \cite{vcadik2013learning} who employed such a strategy for detecting localised distortions in images. The insight is that these metrics each look for slightly different cues within the image, which provides a signal that a simple learner can leverage. Here, we learn four different functions, one for each of the experiments (two tasks and two materials, indoor and outdoor are considered jointly). 



\subsection{Formulation and training}
We seek a function $f_e :\mathbb{R}^2 \mapsto \mathbb{R}$ for experiment $e \in \{1, \ldots, 4\}$ which maps a pair of input images $\vb{I}_a, \vb{I}_b$ to a perception score $\varphi_{a,b}$ (as in \cref{eq:input_data}), such that $f_e(\vb{I}_a,\vb{I}_b) \approx \varphi_{a,b}$:
\begin{equation}
f_e(\vb{I}_a, \vb{I}_b) \equiv \psi_e(\{\ell_k(\vb{I}_1, \vb{I}^*) - \ell_k(\vb{I}_2, \vb{I}^*)\}_{k=1}^K) \,,
\label{eqn:fe}
\end{equation}
where $\vb{I}^*$ is the ground truth image and $\ell_k, k \in \{1, \ldots\, K\}$ is the set of all $K=15$ metrics from \cref{sec:correlation}.  
$\psi_e$ is implemented using an SVR ($\epsilon$-Support Vector Regression \cite{platt1999probabilistic}), with $\epsilon=0.1$ and $C=1$. As shown in \cref{eqn:fe}, the SVR takes as input the set of differences between all $K$ metrics scores computed on the images and their corresponding GT. 
The SVR was selected as the best variant amongst a set of simple learners (see the supplementary material for more details).

To train $\psi_e$, we split our perceptual data into \valeur{\num{20}} training and \valeur{\num{5}} validation scenes and consider the comparisons without replacement between \valeur{\num{5}}/\valeur{\num{3}} indoor/outdoor methods (yielding \valeur{\num{10}}/\valeur{\num{6}} combinations), respectively.
Considering permutations of the input combinations to make $\psi_e$ agnostic to the order of the input images, the resulting dataset contains \valeur{\num{640}} and \valeur{\num{160}} data points for training and validation.

We report the results of our metric on our validation set in \cref{fig:correlation_in_and_out_observers_metrics}, where our proposed metric dominates all other metrics. This demonstrates that existing metrics do capture valuable information and that, while a \emph{single} metric does not accurately match perception in all cases, a \emph{learned combination} of them can approximate it much more precisely.

\subsection{Generalisation to other methods}

We evaluate the generalisation capabilities of our learned metric on other lighting estimation methods in two ways. First, we hold out all perception data of one of the methods (Khan~\etal~\cite{khan2006image} was selected arbitrarily) and retrained $\psi_e$ on the remaining methods. Once the metric is trained, we test it on the unseen (Khan) data, and show results in the ``Ours Holdout'' column of \cref{fig:correlation_in_and_out_observers_metrics}. This is done only for the indoor data as outdoor data is too scarce for such analysis.

Second, we run another psychophysical study with \num{6} observers using the same methodology as in \cref{sec:psychophys_exp}, but employing different indoor lighting estimation methods. This time, we use a re-implementation of EverLight~\cite{Dastjerdi_2023_ICCV} at $1024\times 2048$ resolution, Weber~\etal~\cite{weber2022editable} (used previously but never in paired comparison with the new methods), Garon~\etal~\cite{garon2019fast}, a Stable Diffusion model trained for image outpainting~\cite{rombach2022high}, and a simple baseline where the environment map has constant colour equal to the mean colour of the input image. 
Our metric, trained on all the data points obtained in the main user study, obtains an agreement score for the diffuse/glossy experiment of \valeur{\num{0.786}}/\valeur{\num{0.856}} for task 1 and \valeur{\num{0.787}}/\valeur{\num{0.889}} for task 2.  These agreement scores are superior to what the best standard metrics in \cref{sec:correlation}---VIF, NIQE, RGB Ang. Err., BRISQUE---can provide (\valeur{\num{0.670}}/\valeur{\num{0.809}} and \valeur{\num{0.686}}/\valeur{\num{0.691}}), which validate the generalisation of our proposed metric.  The agreement scores of the best standard metrics are often outliers, with metrics on average performing around chance level.


By releasing all data and code publicly, we hope this learned metric combination will serve as guide for future work in lighting estimation by providing a way to validate its performance in a perceptually meaningful manner.





%% file: sec/7_conc.tex
\section{Discussion}
\label{sec:conc}


This paper explores a new perceptual evaluation framework for lighting estimation algorithms when used for relighting virtual objects into photographs. To do so, it presents a controlled psychophysical study which compares several lighting estimation methods from the recent literature by asking observers to judge scenes rendered with lighting estimates. It then demonstrates that image quality assessment (IQA) metrics typically used to quantitatively evaluate lighting estimation methods seldom coincide with human perception. 

Our analysis emanating from this study provides some key insights. 
Most importantly, human perception is dependent on the task at hand. While IQA metrics generate almost identical results for both our tasks, observer behaviour changes according to whether they must match lighting to a reference, or judge the plausibility of virtual objects in context of a background image. Both scenarios should be considered when evaluating future methods.


To help future lighting estimation approaches in evaluating their performance in a perceptually meaningful way, we propose a learned metric combination which is shown, through generalisation experiments, to accurately predict observers preference on images from unseen methods. 


Despite the novel insights our analysis provides and our proposed learned metric combination, understanding the link between human perception and lighting estimation algorithms is still largely uncharted---our paper is but a first step. Further analysis of how specific algorithmic design choices (e.g., choice of lighting representation) affect human perception is an important future direction to explore. 
%
It is our hope that this work will pave the way for significant research into effective methods for evaluating the lighting estimation problem, methods that will be more tightly tied to human perception. 



\footnotesize{This research was supported by Sentinel North and NSERC grant RGPIN 2020-04799. JVC was supported by Grant PID2021-128178OB-I00 funded by MCIN/AEI/10.13039/501100011033, ERDF ``A way of making Europe'' and by the  Departament de Recerca i Universitats from Generalitat de Catalunya with reference 2021SGR01499.}